\documentclass[10pt,twocolumn,letterpaper]{article}

\usepackage[pagenumbers]{cvpr} % To force page numbers, e.g. for an arXiv version

\usepackage{graphicx}
\usepackage{amsmath}
\usepackage{amssymb}
\usepackage{booktabs}
\usepackage{multirow}

\usepackage[numbers, sort&compress]{natbib}
\setcitestyle{numbers, sort&compress}

\usepackage[belowskip=-7pt,aboveskip=3pt]{caption}

\usepackage[pagebackref, breaklinks, colorlinks]{hyperref}

\usepackage[capitalize]{cleveref}
\crefname{section}{Sec.}{Secs.}
\Crefname{section}{Section}{Sections}
\Crefname{table}{Table}{Tables}
\crefname{table}{Tab.}{Tabs.}
\usepackage{xcolor}

\def\cvprPaperID{1279} % *** Enter the CVPR Paper ID here
\def\confName{CVPR}
\def\confYear{2023}

\newcommand\nnfootnote[1]{%
  \begin{NoHyper}
  \renewcommand\thefootnote{}\footnote{#1}%
  \addtocounter{footnote}{-1}%
  \end{NoHyper}
}

\begin{document}

\title{Cascaded Local Implicit Transformer for Arbitrary-Scale Super-Resolution}

\author{
Hao-Wei Chen\textsuperscript{*,1,2} \ Yu-Syuan Xu\textsuperscript{*,2} \ Min-Fong Hong\textsuperscript{2} \ Yi-Min Tsai\textsuperscript{2} \ Hsien-Kai Kuo\textsuperscript{2} \ Chun-Yi Lee\textsuperscript{1}
\smallskip
\\
\textsuperscript{1}Elsa Lab, National Tsing Hua University~~~ \textsuperscript{2}MediaTek Inc.\\
{\tt\small \{jaroslaw1007, cylee\}@gapp.nthu.edu.tw} \\
{\tt\small \{Yu-Syuan.Xu, romulus.hong, Yi-Min.Tsai, Hsienkai.Kuo \}@mediatek.com}
}

\maketitle
\nnfootnote{*Equal contribution}

%%%%%%%%% ABSTRACT

\begin{abstract}

Implicit neural representation has recently shown a promising ability in representing images with arbitrary resolutions. In this paper, we present a Local Implicit Transformer (LIT), which integrates the attention mechanism and frequency encoding technique into a local implicit image function. We design a cross-scale local attention block to effectively aggregate local features. To further improve representative power, we propose a Cascaded LIT (CLIT) that exploits multi-scale features, along with a cumulative training strategy that gradually increases the upsampling scales during training. We have conducted extensive experiments to validate the effectiveness of these components and analyze various training strategies. The qualitative and quantitative results demonstrate that LIT and CLIT achieve favorable results and outperform the prior works in arbitrary super-resolution tasks. The source codes are accessible at the following repository: \href{https://github.com/jaroslaw1007/CLIT}{https://github.com/jaroslaw1007/CLIT}.

\end{abstract}

%%%%%%%%% INTRODUCTION
\section{Introduction}
\label{sec::introduction}
Single Image Super-Resolution (SISR) is the process of reconstructing high-resolution (HR) images from their corresponding low-resolution (LR) counterparts. SISR has long been recognized as a challenging task in the low-level vision domain due to its ill-posed nature, and has attracted a number of researchers dedicated to this field of study over the past decade~\cite{srcnn, vdsr, fsrcnn, espcn, lapsrn, srresnet, edsr, esrgan, rdn, rcan, san, csnl, ipt, swinir, hat, metasr, liif, ultrasr, ipe, itsrn, lte}. A line of SISR research referred to as `\textit{fixed-scale SR}'~\cite{srcnn, vdsr, fsrcnn, espcn, lapsrn, srresnet, edsr, esrgan, rdn, rcan, san, csnl, ipt, swinir, hat} focuses on extracting feature embeddings from LR images and leveraging these embeddings to upsample images with a predefined factor through learnable deconvolutions~\cite{fsrcnn} or sub-pixel convolutions~\cite{espcn}. Despite their success, many of the proposed approaches necessitate a distinct deep neural network model for each upsampling scale, which is usually restricted to a limited selection of integers~(\eg, $2\times$, $3\times$, $4\times$). Such a limitation constrains the potential applications and deployment options of SISR models. To overcome this limitation, approaches for upsampling LR images in a continuous manner via a single model emerge and attracted considerable attention recently.

\begin{figure}[t]
  \centering
  \includegraphics[width=1\linewidth]{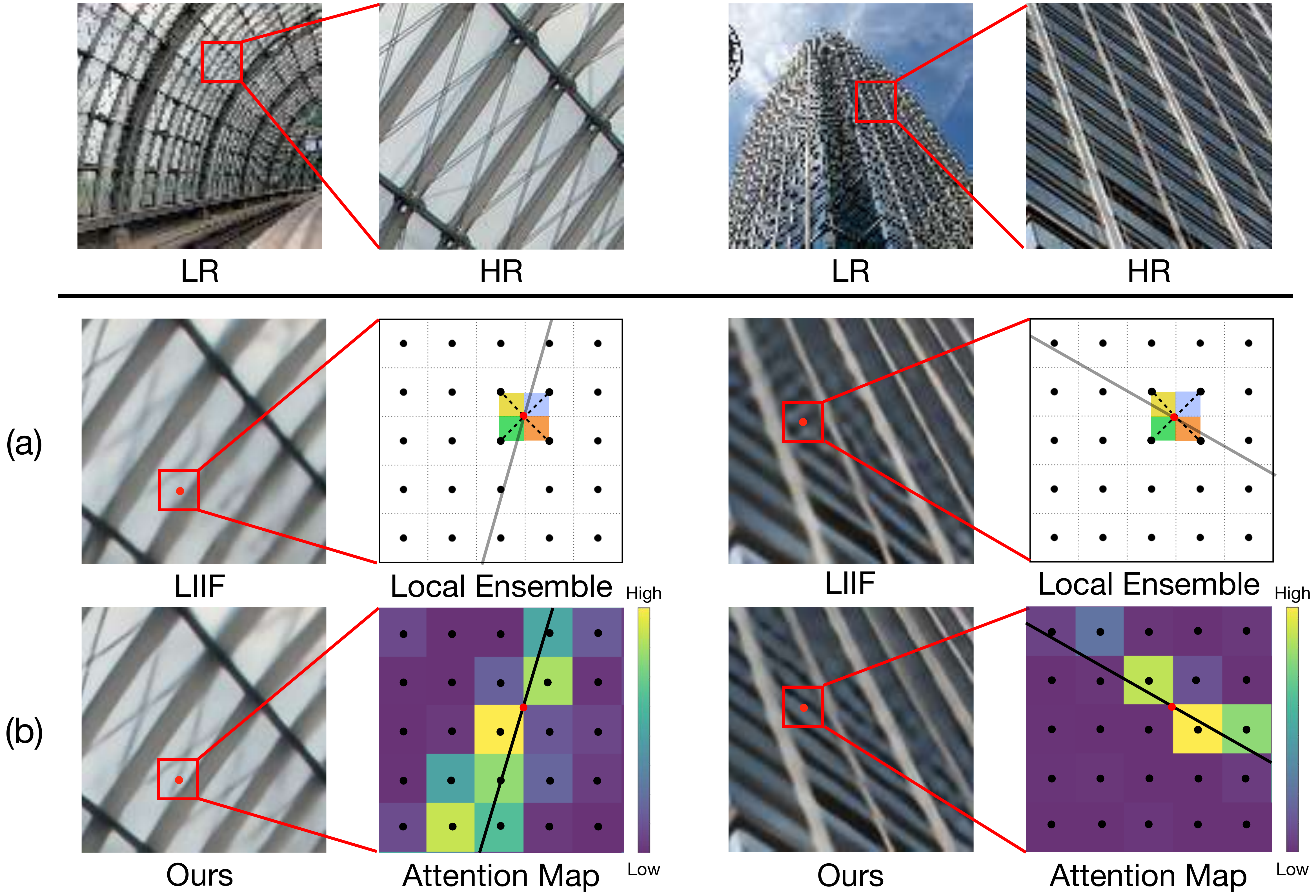}
  \caption{An illustration and comparison of different approaches that take into account nearby pixels for continuous upsampling: (a) the local ensemble method used in~\cite{liif}, and (b) our proposed local attention mechanism.}
  \label{fig:local_concept}
\end{figure}

Over the past few years, arbitrary-scale SR has emerged and attracted considerable attention from researchers~\cite{metasr, liif, ultrasr, ipe, itsrn, lte}.Apart from the pioneering work Meta-SR~\cite{metasr}, recent endeavors~\cite{liif, ultrasr, ipe, itsrn, lte} have achieved arbitrary-scale SR by replacing the upsampling layers commonly adopted by previous approaches with local implicit image functions, and have demonstrated favorable performance. These local implicit functions employ multi-layer perceptrons (MLPs) to map 2D coordinates and corresponding latent representations to RGB values. Fig.~\ref{fig:local_concept} illustrates how different approaches sample latent representations based on the queried coordinates (depicted as the red dots). Fig.~\ref{fig:local_concept}~(a) illustrates the local ensemble technique adopted by contemporary mainstream methods~\cite{liif, ultrasr, ipe, itsrn, lte}. It calculates the RGB value of the queried coordinate by taking the weighted average of those of the surrounding four pixels based on their relative distances to the queried coordinate. This approach, however, does not consider contextual information and relies solely on distance.  For instance, in Fig.~\ref{fig:local_concept}, the queried coordinates are intentionally designed to lie on edges. However, merely calculating the weighted average of pixels fails to reflect the contextual information about the image content, thereby preventing the accurate capture of the necessary features for performing SR. As a result, although pixel distance plays a vital role in SR tasks, it is essential to concentrate more on the contextual information in an image.

In light of the above observations, we propose a Local Implicit Transformer (LIT), which expands the numbers of referenced latent vectors and accounts for the feature correlation in the context by exploiting the attention mechanism~\cite{transformer}. LIT comprises a Cross-Scale Local Attention Block (CSLAB) and a decoder. CSLAB generates attention maps based on the bilinearly interpolated latent vectors at the queried coordinates and the key latent vectors sampled from a grid of coordinates with relative positional bias~\cite{swin, crossformer}. The first and second columns of Fig.~\ref{fig:local_concept}~(b) visualize the attention maps generated by LIT, where the attention areas align closely with the edges. By applying attention maps to feature embeddings, the RGB values of the queried coordinates can be contextually predicted. Then, a decoder is adopted to produce RGB values by taking advantage of the attention feature embedding.

In order to address the issue of diverse scaling factors and achieve arbitrary-scale super-resolution, it is crucial to consider the role of upsampling factors in constructing high-resolution images within the local implicit image function. However, simultaneously training a local implicit image function with a wide range of upsampling factors (e.g., $1\times\sim30\times$) poses significant challenges. As a result, we propose a cumulative training strategy to incrementally enhance the fuction's its representative power. The strategy initially trains the local implicit image function with small upsampling factors and then finetunes it with alternatively sampled small and large ones. Furthermore, we present Cascaded LIT (CLIT) to harness the advantages of multi-scale feature embeddings, complementing missing details and information during one-step upsampling. The combination of the cumulative training strategy and CLIT enables efficient and effective handling of arbitrary-scale SR tasks.

The main contributions of our work are summarized as follows: 
(1) We introduce the LIT architecture, which incorporates the local attention mechanism into arbitrary-scale SR
(2) We further develop a cumulative training strategy and the cascaded framework CLIT to effectively handle large-scale upsampling.
(3) We carry out comprehensive analyses of the performance impacts for LIT and CLIT. The extensive experimental findings demonstrate that the proposed LIT and CLIT are able to yield remarkable or comparable results across a wide range of benchmark datasets.

The paper is organized as follows. Section~\ref{sec::related_works} reviews the related work. Section~\ref{sec::methodology} walks through the proposed LIT and CLIT frameworks and the implementation details. Section~\ref{sec::experimentas} presents the experimental results. Section~\ref{sec::conclusion} concludes.

%%%%%%%%% RELATED_WORKS
\section{Related Work}
\label{sec::related_works}

\paragraph{Implicit neural representation.}
Implicit neural representation is a technique for representing continuous-domain signals via coordinate-based multi-layer perceptrons (MLPs). Its concept has been adopted in various 3D tasks,~\eg, 3D object shape modeling~\cite{chen2019, deepsdf, mescheder2019, genova2020, sal}, 3D scene reconstruction~\cite{srn, jiang2020, chabra2020, peng2020}, and 3D structure rendering~\cite{nerf, niemeyer2020, nsvf, mipnerf}.  For example, NeRF~\cite{nerf} employs implicit neural representation to perform novel view synthesis, which maps coordinates to RGB colors for a specific scene. In the past few years, 2D applications of implicit neural representation have been attempted as well, such as image representation~\cite{klocek2019, siren} and super-resolution~\cite{liif, ultrasr, ipe, lte, itsrn}. Our work is related to a technique called `\textit{local implicit neural representation}'~\cite{liif, lte}, which encodes LR images to feature embeddings such that similar information could be shared within local regions. Such local implicit neural representations are exploited to upscale LR images to HR ones.

\paragraph{Single image super-resolution.}
In the past several years, various deep neural network (DNN) based architectures~\cite{srcnn, vdsr, fsrcnn, espcn, lapsrn, srresnet, edsr, esrgan, rdn, rcan, san, csnl, ipt, swinir, hat} have been proposed for SISR. Among these works, SRCNN~\cite{srcnn} pioneered the use of convolutional neural networks (CNNs) to achieve SISR in an end-to-end manner. It is later followed by several subsequent works that incorporated more complicated model architectures, such as residual blocks~\cite{edsr, srresnet}, dense connections~\cite{rdn, esrgan}, attention based mechanisms~\cite{rcan, san, csnl}, or cascaded frameworks~\cite{lapsrn, csrip, cdpn}, to extract more effective feature representations for SISR. Recently, transformer-based methods~\cite{ipt, swinir, hat} were introduced to SISR and achieved promising performance.

\paragraph{Arbitrary-scale super-resolution.}
As discussed in Section~\ref{sec::introduction}, most of the contemporary SISR works limit their upsampling scales to specific integer values, and are required to train a distinct model for each upsampling scale. To overcome such a limitation, several approaches~\cite{metasr, liif, ultrasr, ipe, itsrn, lte} were proposed to train a unified model for arbitrary upsampling scales. Meta-SR~\cite{metasr} proposed a meta-upscale module for predicting the weights of their convolutional filters from coordinates and scales. The predicted weights are then utilized to perform convolutions to generate HR images. In contrast to Meta-SR, LIIF~\cite{liif} employs an MLP as a local implicit function, which takes a queried coordinate in an HR image, its nearby feature representations extracted from the corresponding LR image, as well as a cell size to predict an RGB value for that coordinate. UltraSR~\cite{ultrasr} and IPE~\cite{ipe} extended LIIF by replacing coordinates with the embedded ones to deal with the spectral bias issue~\cite{nerf, rahaman2019, basri2020, siren, tancik2021} inherent in MLPs. LTE~\cite{lte} further introduced a local texture estimator that transforms coordinates into Fourier domain information to enrich the representational capability of its local implicit function. Different from the above approaches, our proposed methodology exploits a novel local attention mechanism and a cascaded framework to deal with the arbitrary-scale SR. In order to fairly compare with the above approaches, we similarly adopt EDSR~\cite{edsr}, RDN~\cite{rdn} and SwinIR~\cite{swinir} as the encoders for our LIT and CLIT.

%%%%%%%%% METHODOLOGY
\begin{figure*}[t]
    \centering
    \includegraphics[width=1\textwidth]{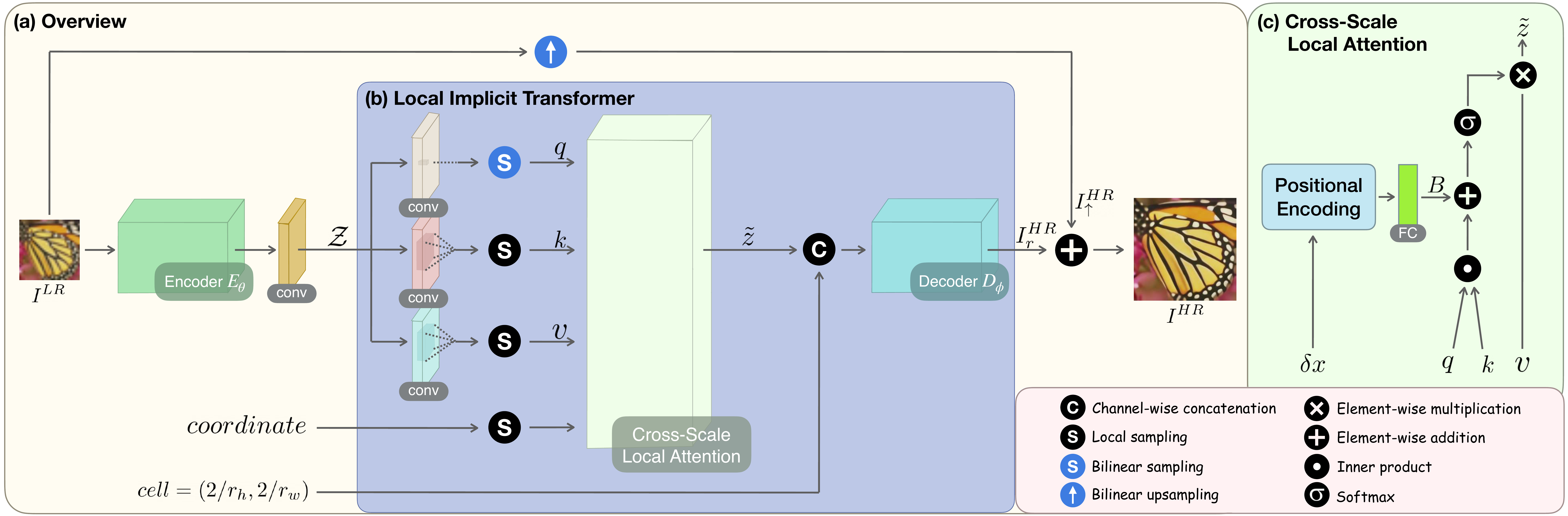}
    \caption{The proposed LIT framework. The local sampling operation samples input embeddings based on a grid of coordinates.}
    \label{fig:architecture}
\end{figure*}

\section{Methodology}
\label{sec::methodology}
In this section, we first provide an overview of the proposed LIT framework, followed by the implementation details of it and its main modules.We then discuss our cumulative training strategy, as well as the framework of CLIT. 

\subsection{Overview of the LIT Framework}
\label{subsec::overview}
LIT is a framework that employs a novel cross-scaled local attention mechanism and a local frequency encoding technique to perform arbitrary-scale SR tasks. Fig.~\ref{fig:architecture}~(a) provides an overview of the proposed framework, which aims at producing an HR image $I^{HR} \in \mathbb{R}^{r_{h}H \times r_{w}W \times 3}$ at 2D HR coordinates $\mathbf{x}^{HR} \in \mathcal{X}$ from a given LR image $I^{LR} \in \mathbb{R}^{H \times W \times 3}$ at 2D LR coordinates $\mathbf{x}^{LR} \in \mathcal{X}$ based on an arbitrary upsampling scale $\mathbf{r} = \left \{r_{h}, r_{w} \right \}$, where $\mathcal{X}$ is the 2D coordinate space that is used to represent an image in the continuous domain. An encoder $E_{\theta}$ first extracts a feature embedding $\mathcal{Z} \in \mathbb{R}^{H \times W \times C}$ from $I^{LR}$.  The extracted $\mathcal{Z}$ is then forwarded into LIT along with the 2D coordinates of $I^{HR}$ and a cell = $(2/s_{h}, 2/s_{w})$ to generate the RGB values of a residual image $I^{HR}_{r} \in \mathbb{R}^{r_{h}H \times r_{w}W \times 3}$ in a pixel-wise fashion. Lastly, the residual image $I^{HR}_{r}$ is combined with a bilinearly upsampled image $I^{HR}_{\uparrow}\in \mathbb{R}^{r_{h}H \times r_{w}W \times 3}$ via element-wise addition to derive the final HR image $I^{HR}$. 

\subsection{Local Implicit Transformer}
\label{subsec::lit}
\begin{figure}[t]
    \centering
    \includegraphics[width=1\linewidth]{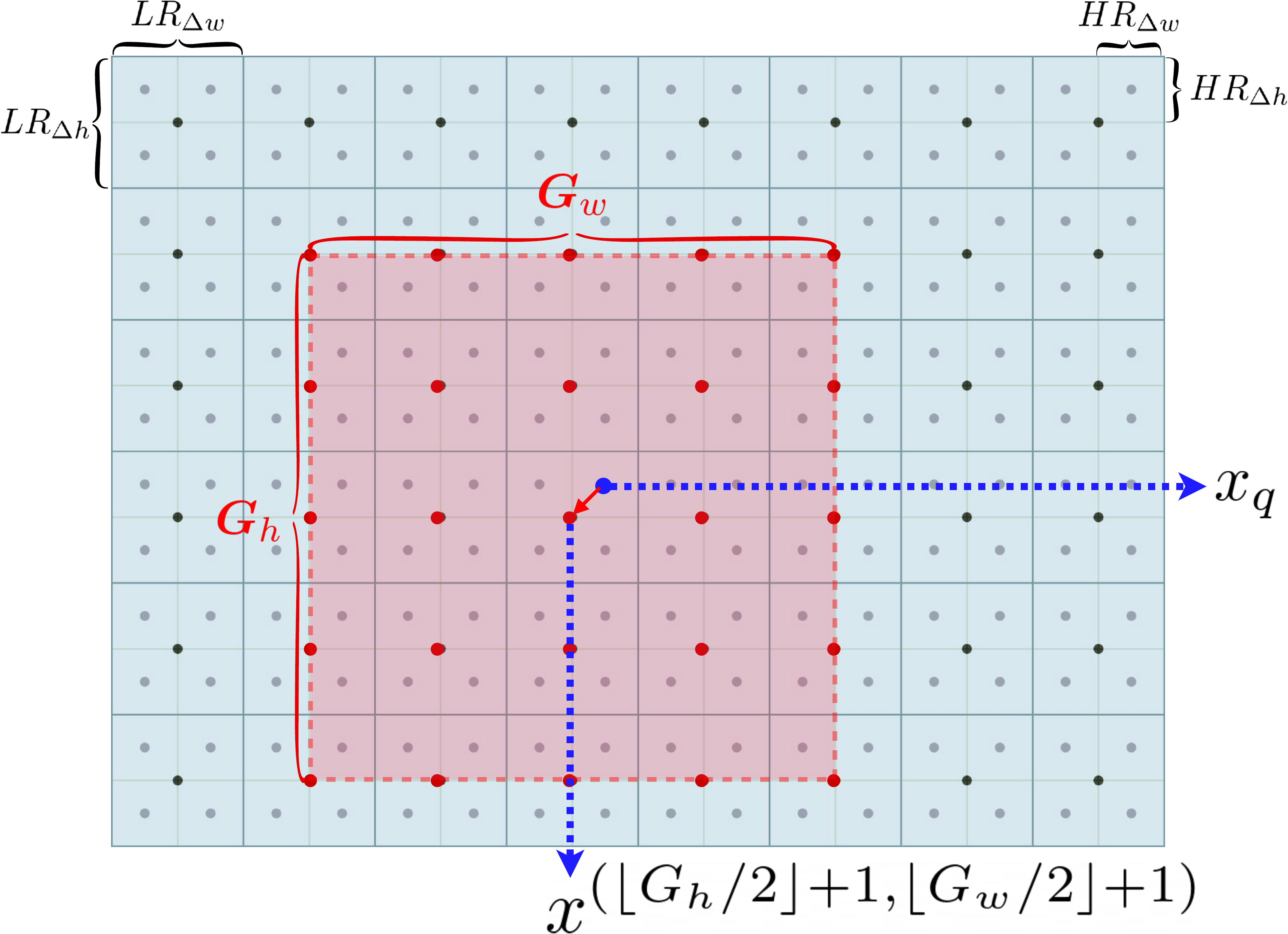}
    \caption{An illustration of the proposed local coordinate sampling scheme. The \textcolor{red}{red} rectangular region, outlined by a dashed line, represents the local grid with dimensions $G_{h} \times G_{w}$. The \textcolor{red}{red dots} indicate the sampled local LR coordinates, while the \textcolor{gray}{gray dots} correspond to the HR coordinates. The grid dimensions $LR_{\Delta_{h, w}}$ and $HR_{\Delta_{h, w}}$ represent the unit sizes of pixels in LR and HR images, respectively.}
    \label{fig:local_coord}
\end{figure}

LIT is developed for mapping any 2D coordinate in the continuous image domain to an RGB color. As highlighted in Fig.~\ref{fig:architecture}~(b), it is composed of a cross-scale local attention block (CSLAB) and a decoder $D_{\phi}$ paraterizerized by $\phi$. CSLAB is responsible for estimating a local latent embedding $\tilde{z} \in \mathbb{R}^{G_{h}G_{w} \times C}$, where $G_{h}$ and $G_{w}$ denote the height and width of local grids employed for performing local coordinate sampling, as depicted in Fig.~\ref{fig:local_coord}. On the other hand, $D_{\phi}$ utilizes these embeddings along with the provided cell to generate $I^{HR}_{r}$. More specifically, LIT first projects $\mathcal{Z}$ using four separate convolutional layers to obtain three latent embeddings, corresponding to query $q$, key $k$, and value $v$. Based on a queried HR coordinate $x_{q} \in \mathbf{x}^{HR}$, CSLAB estimate $\tilde{z}$ as follows:
\begin{equation}
    \tilde{z} = CSLAB(\delta x, q, k, v), 
\label{equation1}
\end{equation}
\begin{equation}
    \delta \mathbf{x} =  \left \{x_{q} - x^{(i, j)} \right \}_{i \in \left \{1, 2, ..., G_{h}\right \}, j \in \left \{1, 2, ..., G_{w}\right \}}, 
\label{equation3}
\end{equation}
where $x^{(i, j)} \in \mathbf{x}^{LR}$ denotes an LR coordinate in the local grid indexed by $(i, j)$, and $\delta \mathbf{x}$ represents the set of local relative coordinates defined by Eq.~(\ref{equation3}). The local grid is sampled in a manner that positions its center $x^{(\left \lfloor G_{h}/2 \right \rfloor +1, \left \lfloor G_{w}/2 \right \rfloor +1)}$ at the LR coordinate closest to $x_{q}$. The query latent vector $q \in \mathbb{R}^{1 \times C}$ at the HR coordinate $x_{q}$ is computed by bilinear interpolation, while the remainder of the local latent embeddings $k \in \mathbb{R}^{G_{h}G_{w} \times C}$ and $v \in \mathbb{R}^{G_{h}G_{w} \times C}$ are sampled at the local LR coordinates $\mathbf{x} = \{x^{i,j}\}_{i \in \left \{1, 2, ..., G_{h}\right \}, j \in \left \{1, 2, ..., G_{w}\right \}}$. With the local latent embedding $\tilde{z}$, the function of $D_{\phi}$ is formulated as follows:
\begin{equation}
    I^{r}(x_{q}) =  D_{\phi}(\tilde{z}, c),
\label{equation4}
\end{equation}
where $I^r(x_{q})$ is the predicted RGB value at the queried coordinate $x_{q}$, and $c = \{HR_{\Delta h}, HR_{\Delta w}\}$ denotes the cell that represents the height and width of a pixel in an HR image, as illustrated in Fig.~\ref{fig:local_coord}. $D_{\phi}$ is implemented as a five-layer MLP utilizing Gaussian Error Linear Unit (GELU) activation~\cite{gelu}, and is employed consistently across all images.

\paragraph{Cross-scale local attention block.}
LIT exploits CSLAB to perform a local attention mechanism over a local grid to generate a local latent embedding $\tilde{z}$ for each HR coordinate, as illustrated in Fig.~\ref{fig:architecture}~(c). CSLAB first calculates the inner product of $q$ and $k$, adds the relative positional bias $B$ to the result, and obtains an attention matrix. This attention matrix is subsequently normalized by a Softmax operation to produce a local attention map. Finally, CSLAB performs element-wise multiplication of $v$ and the local attention map to derive $\tilde{z}$. The overall procedure is formulated as follows:
\begin{equation}
    \tilde{z} =  softmax(\frac{qk^{\top}}{\sqrt{C}} + B) \times v ,
\label{equation5}
\end{equation}
\begin{equation}
    B =  FC(\gamma (\delta \mathbf{x})) ,
\label{equation6}
\end{equation}
\begin{equation}
\begin{aligned}
    \gamma(\delta \mathbf{x}) &= [\sin{(2^{0} \delta \mathbf{x})}, \cos{(2^{0} \delta \mathbf{x})}, ... \\
    & ,\sin{(2^{L-1} \delta \mathbf{x})}, \cos{(2^{L-1}  \delta \mathbf{x})}],
\end{aligned}
\label{equation7}
\end{equation}
where $C$ denotes the channel dimension of the local key latent embedding $k$, $FC$ represents a fully-connected layer, $\gamma$ is the positional encoding function, and $L$ is a hyperparameter. In this work, $L$ is set to $10$, and the multi-head attention mechanism adopted is formulated in Eq.~(\ref{equation8}) as follows:
\begin{equation}
    \tilde{z} =  concat(softmax(\frac{q_{i}k_{i}^{\top}}{\sqrt{C/H}} + B_{i}) \times v_{i}) ,
\label{equation8}
\end{equation}
where $H$ is the number of attention heads and $i \in [1,..., H]$.

\subsection{Cumulative Training Strategy}
\label{subsec::cumulative_training_strategy}
In this section, we discuss our proposed  cumulative training strategy, which is developed for enhancing the performance of arbitrary-scale SR.  The cumulative training strategy focuses on the schedule of the cell sizes selected during the training phase, as ``cell decoding" has been recognized as an essential input to a local implicit image function~\cite{liif}. Recent studies~\cite{liif,ipe} have observed that the effect of cell decoding on the performance of arbitrary-scale SR is prominent for in-distribution upsampling, but degrades significantly for out-of-distribution large-scale upsampling. Such trends are demonstrated in Table~3 of~\cite{liif} and Table~4 of~\cite{ipe}. 
The authors in~\cite{lte} also constrained their cell sizes to in-distribution ranges to mitigate the negative impact during evaluation. To overcome the degradation issue for out-of-distribution cell sizes, incorporating large cell sizes during training appears to be a promising solution. However, simply training the local implicit image function with a diverse range of cell sizes at once leads to a performance drop. According to our experimental observations presented in Section~\ref{subsec::ablation_studies}, we find that training the local implicit image function by alternatively switching between large and small cells offers positive impacts on the performance. Based on the above observations, we propose a cumulative training strategy which first trains the local implicit image function with large cell sizes, and finetunes it with the alternative training strategy to improve the performance on different upsampling scales. More quantitative results of our training strategies can be found in Section~\ref{subsec::ablation_studies}.

\begin{figure}[t]
    \centering
    \includegraphics[width=1\linewidth]{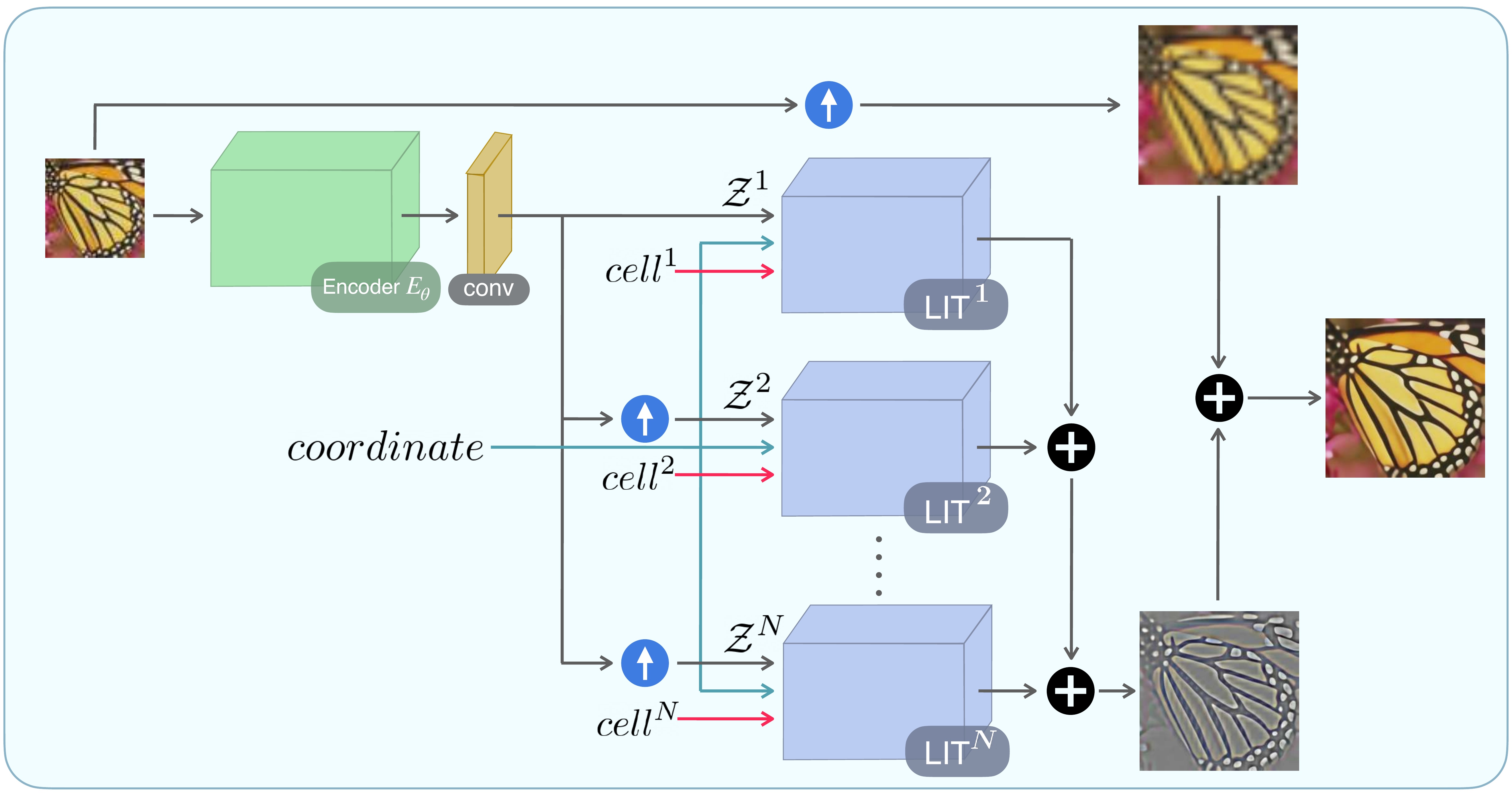}
    \caption{An overview of the proposed CLIT framework.}
    \label{fig:cascaded_lit}
\end{figure}

\begin{table*}[t]
\centering
\footnotesize
\resizebox{1\textwidth}{!}{
\begin{tabular}{l|cccccccc}
Method                                  & $\times 2$              & $\times 3$              & $\times 4$              & $\times 6$              & $\times 12$             & $\times 18$             & $\times 24$             & $\times 30$              \\ \hline\hline
Bicubic~\cite{edsr}                     & 31.01                   & 28.22                   & 26.66                   & 24.82                   & 22.27                   & 21.00                   & 20.19                   & 19.59                    \\ 
\hline
EDSR-baseline~\cite{edsr}               & 34.55                   & 30.90                   & 28.94                   & -                       & -                       & -                       & -                       & -                        \\
EDSR-baseline-Meta-SR~\cite{liif,metasr}& 34.64                   & 30.93                   & 28.92                   & 26.61                   & 23.55                   & 22.03                   & 21.06                   & 20.37                    \\
EDSR-baseline-LIIF~\cite{liif}          & 34.67                   & 30.96                   & 29.00                   & 26.75                   & 23.71                   & 22.17                   & 21.18                   & 20.48                    \\
EDSR-baseline-UltraSR~\cite{ultrasr}    & 34.69                   & \textcolor{blue}{31.02} & 29.05                   & \textcolor{blue}{26.81} & 23.75                   & 22.21                   & 21.21                   & 20.51                    \\
EDSR-baseline-IPE~\cite{ipe}            & \textcolor{blue}{34.72} & 31.01                   & 29.04                   & 26.79                   & 23.75                   & 22.21                   & 21.22                   & 20.51                    \\
EDSR-baseline-LTE~\cite{lte}            & \textcolor{blue}{34.72} & \textcolor{blue}{31.02} & \textcolor{blue}{29.04} & \textcolor{blue}{26.81} & \textcolor{blue}{23.78} & \textcolor{blue}{22.23} & \textcolor{blue}{21.24} & \textcolor{blue}{20.53}  \\
EDSR-baseline-CLIT (Ours)               & \textcolor{red}{34.82}  & \textcolor{red}{31.14}  & \textcolor{red}{29.17}  & \textcolor{red}{26.93}  & \textcolor{red}{23.85}  & \textcolor{red}{22.30}  & \textcolor{red}{21.27}  & \textcolor{red}{20.54}   \\ \hline
RDN-Meta-SR~\cite{liif,metasr}          & 35.00                   & 31.27                   & 29.25                   & 26.88                   & 23.73                   & 22.18                   & 21.17                   & 20.47                    \\
RDN-LIIF~\cite{liif}                    & 34.99                   & 31.26                   & 29.27                   & 26.99                   & 23.89                   & 22.34                   & 21.31                   & 20.59                    \\
RDN-UltraSR~\cite{ultrasr}              & 35.00                   & 31.30                   & 29.32                   & 27.03                   & 23.73                   & 22.36                   & 21.33                   & 20.61                    \\
RDN-IPE~\cite{ipe}                      & \textcolor{blue}{35.04} & \textcolor{blue}{31.32} & 29.32                   & \textcolor{blue}{27.04} & 23.93                   & 22.38                   & 21.34                   & 20.63                    \\
RDN-LTE~\cite{lte}                      & \textcolor{blue}{35.04} & \textcolor{blue}{31.32} & \textcolor{blue}{29.33} & \textcolor{blue}{27.04} & \textcolor{blue}{23.95} & \textcolor{blue}{22.40} & \textcolor{blue}{21.36} & \textcolor{red}{20.64}   \\
RDN-CLIT (Ours)                         & \textcolor{red}{35.10}  & \textcolor{red}{31.38}  & \textcolor{red}{29.40}  & \textcolor{red}{27.12}  & \textcolor{red}{24.01}  & \textcolor{red}{22.45}  & \textcolor{red}{21.38}  & \textcolor{red}{20.64}   \\ \hline
SwinIR-MetaSR~\cite{lte,metasr}         & 35.15                   & 31.40                   & 29.33                   & 26.94                   & 23.80                   & 22.26                   & 21.26                   & 20.54                    \\
SwinIR-LIIF~\cite{lte,liif}             & 35.17                   & 31.46                   & 29.46                   & 27.15                   & 24.02                   & 22.43                   & 21.40                   & 20.67                    \\
SwinIR-LTE~\cite{lte}                   & \textcolor{blue}{35.24} & \textcolor{blue}{31.50} & \textcolor{blue}{29.51} & \textcolor{blue}{27.20} & \textcolor{blue}{24.09} & \textcolor{blue}{22.50} & \textcolor{blue}{21.47} & \textcolor{red}{20.73}   \\
SwinIR-CLIT (Ours)                      & \textcolor{red}{35.31}  & \textcolor{red}{31.57}  & \textcolor{red}{29.56}  & \textcolor{red}{27.27}  & \textcolor{red}{24.14}  & \textcolor{red}{22.53}  & \textcolor{red}{21.48}  & \textcolor{blue}{20.71}
\end{tabular}}
\caption{The average PSNR~(dB) on the DIV2K~\cite{div2k} validation set. The results are obtained from the original manuscripts~\cite{liif,ultrasr,ipe,lte}. The best and second-best performing results are highlighted by the \textcolor{red}{red} and \textcolor{blue}{blue} colors, respectively.}
\label{table:div2k_results}
\vspace{-10pt}
\end{table*}

\subsection{Cascaded Local Implicit Transformer}
\label{subsec::clit}
Previous works~\cite{liif, lte} typically address arbitrary-scale SR through a single step of upsampling. 
However, one-step upsampling struggles to reconstruct an HR image when the upsampling scale is large~\cite{lapsrn, csrip, cdpn}. In light of this, we propose a cascaded upsampling strategy, called ``\textit{Cascaded LIT (CLIT)}" to predict residual images from multi-scale feature embeddings. In an $N$-branched CLIT, the multi-scale feature embeddings $\mathcal{Z}^{1}, \mathcal{Z}^{2}, ..., \mathcal{Z}^{N}$ are derived as follows:
\begin{equation}
\begin{aligned}
    &\mathcal{Z}^{N} = \mathcal{Z}\uparrow_{s^{1} \times s^{2} \times ... \times s^{N-1}}, \\
    &\mbox{where } s^{1} = 1 \mbox{ and } s \in \mathbf{s},
\label{equation9}
\end{aligned}
\end{equation}
where $\uparrow$ is a bilinear upsampling function and $\mathbf{s}$ is a set of scaling factors, which are configurable hypermeters. For a branch $i, i\in[1,...,N]$, LIT$^{i}$ estimates the residual image $I^{i}_{r}$ from the feature embedding $\mathcal{Z}^{i}$ with the coordinate and the corresponding cells. Lastly, the final HR image $I^{HR}\in \mathbb{R}^{r_{h}H \times r_{w}W \times 3}$ can be estimated as the following equation:
\begin{equation}
    I^{HR} =  \lambda^{N-1} I^{1}_{r} + \lambda^{N-2} I^{2}_{r} + ... + \lambda^{0} I^{N}_{r} + I^{HR}_{\uparrow}.
\label{equation10}
\end{equation}
where $\lambda$ is a discount factor, with a default value of $0.75$. During the training phase, CLIT is trained using the proposed cumulative training strategy.  Initially, LIT$^{1}$ is trained with the strategy employed by~\cite{liif}. Subsequently, LIT$^{1}$ is fine-tuned and LIT$^{2}$ is initialized by applying the alternative training strategy. By incrementally incorporating LITs into CLIT, the performance is progressively enhanced. The details of the training strategies are specified in Section~\ref{sec::experimentas}.

%%%%%%%%% EXPERIMENTAL_RESULTS
\begin{table*}[t]
\centering
\footnotesize
\resizebox{1\linewidth}{!}{
\begin{tabular}{l|ccccc|ccccc|ccccc|ccccc}
\multirow{2}{*}{Method} & \multicolumn{5}{c|}{Set5}                                                                                                       & \multicolumn{5}{c|}{Set14}                                                                                                      & \multicolumn{5}{c|}{B100}                                                                                                       & \multicolumn{5}{c}{Urban100}                                                                                                     \\
                        & $\times 2$              & $\times 3$              & $\times 4$              & $\times 6$              & $\times 8$              & $\times 2$              & $\times 3$              & $\times 4$              & $\times 6$              & $\times 8$              & $\times 2$              & $\times 3$              & $\times 4$              & $\times 6$              & $\times 8$              & $\times 2$              & $\times 3$              & $\times 4$              & $\times 6$              & $\times 8$               \\ 
\hline\hline
RDN~\cite{rdn}                      & \textcolor{blue}{38.24} & 34.71                   & 32.47                   & -                       & -                       & 34.01                   & 30.57                   & 28.81                   & -                       & -                       & 32.34                   & 29.26                   & 27.72                   & -                       & -                       & 32.89                   & 28.80                   & 26.61                   & -                       & -                        \\
RDN-Meta-SR~\cite{liif, metasr}     & 38.22                   & 34.63                   & 32.38                   & 29.04                   & 26.96                   & 33.98                   & 30.54                   & 28.78                   & 26.51                   & 24.97                   & 32.33                   & 29.26                   & 27.71                   & 25.90                   & 24.83                   & 32.92                   & 28.82                   & 26.55                   & 23.99                   & 22.59                    \\
RDN-LIIF~\cite{liif}                & 38.17                   & 34.68                   & 32.50                   & 29.15                   & 27.14                   & 33.97                   & 30.53                   & 28.80                   & 26.64                   & 25.15                   & 32.32                   & 29.26                   & 27.74                   & 25.98                   & 24.91                   & 32.87                   & 28.82                   & 26.68                   & 24.20                   & 22.79                    \\
RDN-UltraSR~\cite{ultrasr}          & 38.21                   & 34.67                   & 32.49                   & \textcolor{blue}{29.33} & 27.24                   & 33.97                   & \textcolor{blue}{30.59} & 28.86                   & 26.69                   & \textcolor{blue}{25.25} & 32.35                   & 29.29                   & \textcolor{blue}{27.77} & \textcolor{blue}{26.01} & \textcolor{blue}{24.96} & 32.97                   & 28.92                   & 26.78                   & \textcolor{blue}{24.30} & 22.87                    \\
RDN-IPE~\cite{ipe}                  & 38.11                   & 34.68                   & 32.51                   & 29.25                   & 27.22                   & 33.94                   & 30.47                   & 28.75                   & 26.58                   & 25.09                   & 32.31                   & 29.28                   & 27.76                   & 26.00                   & 24.93                   & 32.97                   & 28.82                   & 26.76                   & 24.26                   & 22.87                    \\
RDN-LTE~\cite{lte}                  & 38.23                   & \textcolor{blue}{34.72} & \textcolor{blue}{32.61} & 29.32                   & \textcolor{blue}{27.26} & \textcolor{blue}{34.09} & 30.58                   & \textcolor{blue}{28.88} & \textcolor{blue}{26.71} & 25.16                   & \textcolor{blue}{32.36} & \textcolor{blue}{29.30} & \textcolor{blue}{27.77} & \textcolor{blue}{26.01} & 24.95                   & \textcolor{blue}{33.04} & \textcolor{blue}{28.97} & \textcolor{blue}{26.81} & 24.28                   & \textcolor{blue}{22.88}  \\
RDN-CLIT (Ours)                     & \textcolor{red}{38.26}  & \textcolor{red}{34.80}  & \textcolor{red}{32.69}  & \textcolor{red}{29.39}  & \textcolor{red}{27.34}  & \textcolor{red}{34.21}  & \textcolor{red}{30.66}  & \textcolor{red}{28.98}  & \textcolor{red}{26.83}  & \textcolor{red}{25.35}  & \textcolor{red}{32.39}  & \textcolor{red}{29.34}  & \textcolor{red}{27.82}  & \textcolor{red}{26.07}  & \textcolor{red}{25.00}  & \textcolor{red}{33.13}  & \textcolor{red}{29.04}  & \textcolor{red}{26.91}  & \textcolor{red}{24.43}  & \textcolor{red}{23.03}   \\ 
\hline
SwinIR~\cite{swinir}                & \textcolor{blue}{38.35} & \textcolor{blue}{34.89} & 32.72                   & -                       & -                       & 34.14                   & 30.77                   & 28.94                   & -                       & -                       & \textcolor{blue}{32.44} & 29.37                   & 27.83                   & -                       & -                       & 33.40                   & 29.29                   & 27.07                   & -                       & -                        \\
SwinIR-MetaSR~\cite{lte, metasr}    & 38.26                   & 34.77                   & 32.47                   & 29.09                   & 27.02                   & 34.14                   & 30.66                   & 28.85                   & 26.58                   & 25.09                   & 32.39                   & 29.31                   & 27.75                   & 25.94                   & 24.86                   & 33.29                   & 29.12                   & 26.76                   & 24.16                   & 22.75                    \\
SwinIR-LIIF~\cite{liif}             & 38.28                   & 34.87                   & 32.73                   & 29.46                   & \textcolor{blue}{27.36} & 34.14                   & 30.75                   & 28.98                   & 26.82                   & 25.34                   & 32.39                   & 29.34                   & 27.84                   & 26.07                   & 25.01                   & 33.36                   & 29.33                   & 27.15                   & 24.59                   & 23.14                    \\
SwinIR-LTE~\cite{lte}               & 38.33                   & \textcolor{blue}{34.89} & \textcolor{blue}{32.81} & \textcolor{blue}{29.50} & 27.35                   & \textcolor{blue}{34.25} & \textcolor{blue}{30.80} & \textcolor{blue}{29.06} & \textcolor{blue}{26.86} & \textcolor{blue}{25.42} & \textcolor{blue}{32.44} & \textcolor{blue}{29.39} & \textcolor{blue}{27.86} & \textcolor{blue}{26.09} & \textcolor{blue}{25.03} & \textcolor{blue}{33.50} & \textcolor{blue}{29.41} & \textcolor{blue}{27.24} & \textcolor{blue}{24.62} & \textcolor{blue}{23.17}  \\
SwinIR-CLIT (Ours)                  & \textcolor{red}{38.40}  & \textcolor{red}{34.96}  & \textcolor{red}{32.88}  & \textcolor{red}{29.69}  & \textcolor{red}{27.56}  & \textcolor{red}{34.31}  & \textcolor{red}{30.85}  & \textcolor{red}{29.09}  & \textcolor{red}{26.97}  & \textcolor{red}{25.47}  & \textcolor{red}{32.49}  & \textcolor{red}{29.43}  & \textcolor{red}{27.92}  & \textcolor{red}{26.15}  & \textcolor{red}{25.08}  & \textcolor{red}{33.62}  & \textcolor{red}{29.50}  & \textcolor{red}{27.28}  & \textcolor{red}{24.76}  & \textcolor{red}{23.30}      
\end{tabular}}
\caption{The average PSNR~(dB) on Set5~\cite{set5}, Set14~\cite{set14}, B100~\cite{b100}, and Urban100~\cite{urban100}. The results are obtained from the original manuscripts~\cite{liif,ultrasr,ipe,lte}. The best and second-best performing results are highlighted by the \textcolor{red}{red} and \textcolor{blue}{blue} colors, respectively.}
\label{table:dataset_results}
\end{table*}

\begin{figure*}[t]
    \centering
    \includegraphics[width=0.9\textwidth]{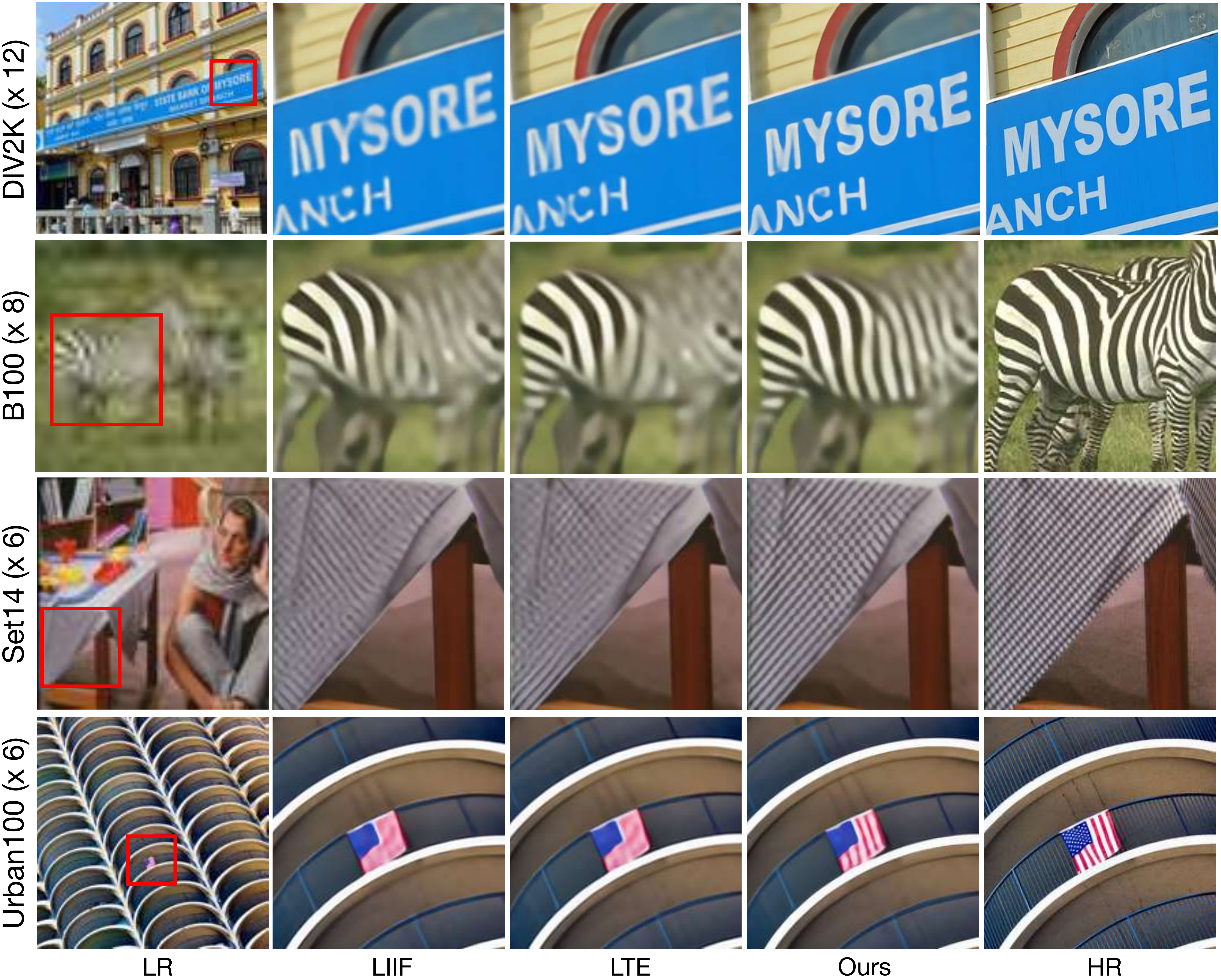}
    \caption{The qualitative results of LIIF~\cite{liif}, LTE~\cite{lte}, and our CLIT with using RDN~\cite{rdn} as the encoder.}
    \label{fig:qualitative_results_fixed}
    \vspace{-10pt}
\end{figure*}

\begin{figure*}[t]
    \centering
    \includegraphics[width=0.9\textwidth]{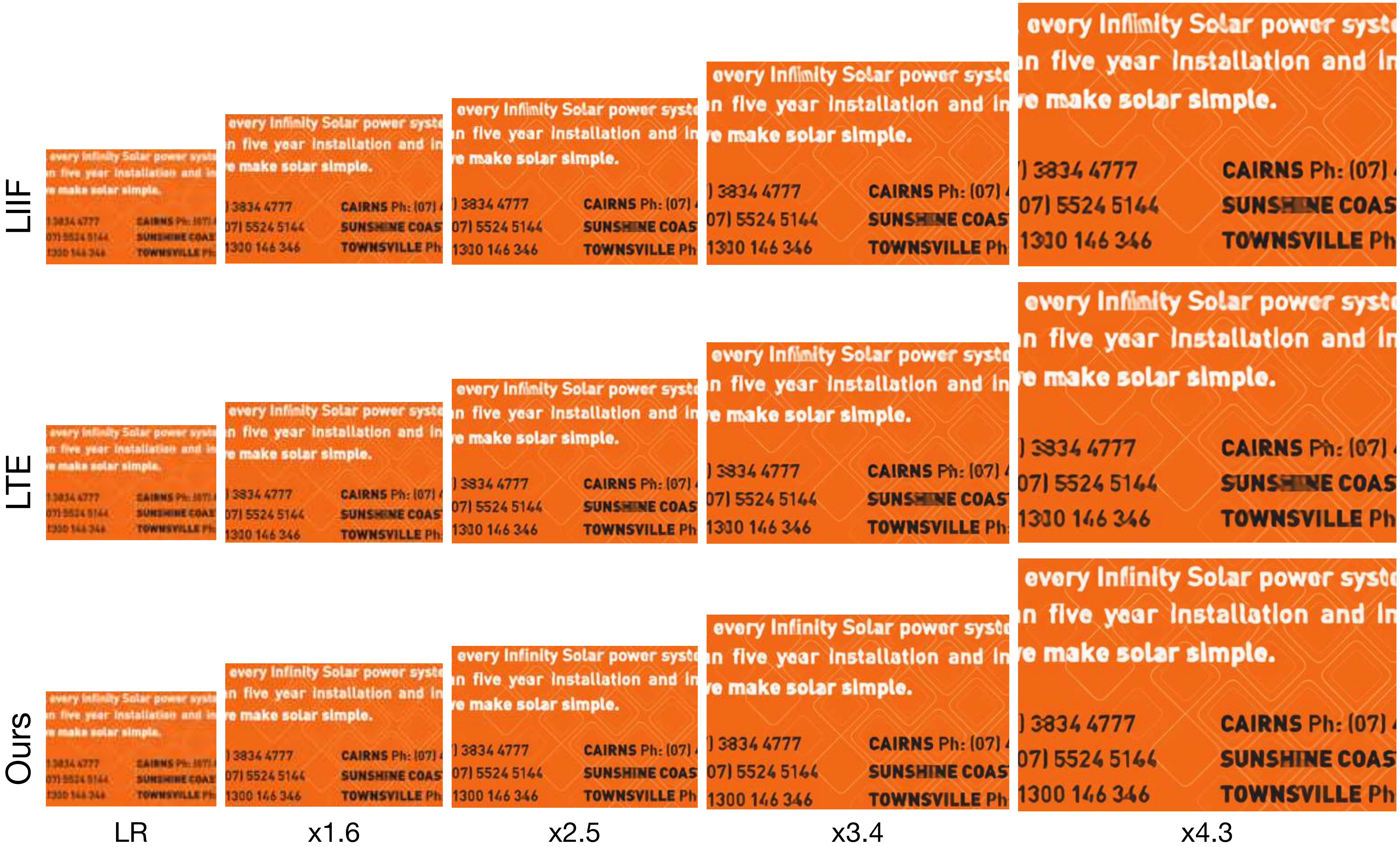}
    \caption{The qualitative results of LIIF~\cite{liif}, LTE~\cite{lte}, and our CLIT using RDN~\cite{rdn} as the encoder and non-integer upsampling scales.}
    \label{fig:qualitative_results_float}
    \vspace{-10pt}
\end{figure*}

\section{Experimental Results}
\label{sec::experimentas}
In this section, we present the experimental results and discuss their implications. We begin with a brief introduction to our experimental setup in Section~\ref{subsec::experimental_setup}. Following that, we evaluate our CLIT with different datasets in Section~\ref{subsec::validation_of_clit}.
The, Section~\ref{subsec::visuallizing_local_attention_maps} showcases the learned attention maps. Finally, ablation studies for various configurations of the proposed CLIT and LIT are compared in Section~\ref{subsec::ablation_studies}.

\subsection{Experimental Setup}
\label{subsec::experimental_setup}

\paragraph{Dataset.}
We use the DIV2K dataset~\cite{div2k} for network training. It consists of $1,000$ images in 2K resolutions and provides low-resolution counterparts with down-sampling scales, $\times 2, \times 3, \times 4$, which are generated by the bicubic interpolation method. On the other hand, we also evaluate the performance on the validation set of DIV2K~\cite{div2k}, Set5~\cite{set5}, Set14~\cite{set14}, B100~\cite{b100} and Urban100~\cite{urban100} in terms of peak signal-to-noise (PSNR) values.

\paragraph{Training details}
During training, we feed batches of size $48 \times 48$ low-resolution images into the framework, following the prior works~\cite{edsr}. For each batch, a single upsampling scale is sampled from a uniform distribution $r \sim \mathcal{U}(1, 4)$. With single upsampling scales $s$, a batch of HR images is cropped into patches of size $48r \times 48r$, while the corresponding LR images are cropped into patches of $48 \times 48$. The patches are augmented by randomly horizontal flipping, vertical flipping, and $90^{\circ}$ rotating. Then, we sample $48^{2}$ pixels (coordinate-RGB pairs) on each HR patch as the ground-truths. We set the batch size to $32$ and use the Adam optimizer~\cite{adam} together with L1 loss for training. We train LIT for $1,000$ epochs, and the learning rate is initialized at $1e^{-4}$ and decayed by a factor 0.5 at [200, 400, 600, 800] epochs. For cumulative training of CLIT, as discussed in Section~\ref{subsec::clit}, we sample $N$ scale factors $\{s^{1}, s^{2}, ..., s^{N}\}$ from the distribution $\mathcal{U}(1, 4)$ according the number of LITs $N$ in the train step. The total upsampling scale $r = s^{1} \times s^{2} ... \times s^{N}$ is the product of all scale factors. If $48r \times 48r$ is greater than the whole HR image, we clip the scale factor of stage 1. To train $N$ LITs, we fine-tune the model for $500 \times N$ epochs, and the learning rate is initialized at $1e^{-4}$ and decayed by a factor 0.5 at [$100 \times N$, $200 \times N$, $300 \times N$, $400 \times N$] epochs.

\subsection{Validation of CLIT}
\label{subsec::validation_of_clit}

\paragraph{Quantitative results.}
We first compare our proposed CLIT to existing local implicit neural representation methods for arbitrary-scale SR, including LIIF~\cite{liif}, UltraSR~\cite{ultrasr}, IPE~\cite{ipe}, and LTE~\cite{lte}. Table~\ref{table:div2k_results} summarizes the quantitative results in terms of PSNR(dB) on DIV2k~\cite{div2k}. As shown in Table~\ref{table:div2k_results}, CLIT achieves the best performance when EDSR-Baseline~\cite{edsr}, RDN~\cite{rdn} are used as the encoders. These results demonstrate the advantage of CLIT.

Table~\ref{table:dataset_results} compares CLIT to prior works~\cite{liif,ultrasr,ipe,lte} on widely-used datasets, including Set5~\cite{set5}, Set14~\cite{set14}, B100~\cite{b100}, and Urban100~\cite{urban100}, with RDN and SwinIR. Note that RDN~\cite{rdn} is trained and evaluated for specific upsampling scales. CLIT outperforms those existing methods in most cases across all datasets and scales, even achieving a 0.20dB PSNR improvement on Set5 for the $\times 8$ scale.

\vspace{-10pt}
\paragraph{Qualitative results.}
Fig.~\ref{fig:qualitative_results_fixed} compares the qualitative results of CLIT with the baseline methods LIIF~\cite{liif} and LTE~\cite{lte}, on various dataset DIV2k~\cite{div2k}, Set14~\cite{set14}, B100~\cite{b100}, and Urban100~\cite{urban100} with variant upsampling scales. The official codes provided by them were used to produce the results. In the first row, the SR visualization results with a $\times12$ upsampling scale are depicted.  It can be observed that both LIIF and LTE struggle to reconstruct the letters continuously, whereas the CLIT result demonstrates continuity for these alphabets, particularly for 'M' and 'S'. In the second row, the stripes on the zebra appear blurry in the LIIF and LTE results,  whereas they are more distinct in the CLIT result. In the third row, the cross pattern on the tablecloth is not clearly rendered by LIIF and LTE.  In contrast, CLIT is able to successfully generate clean crosses and sharp lines. In the fourth row, despite the blurriness of the original LR image, CLIT captures the texture and produces straight lines on the flag, showcasing its effectiveness.

Fig.~\ref{fig:qualitative_results_float} presents the results of text image enhancement employing progressively increasing non-integer upsampling factors, alongside a comparison with earlier methods, LIIF \cite{liif} and LTE \cite{lte}. The input image first undergoes a downscaling process by a factor of $2.2$ to generate an LR image. Subsequently, the LR image is then upscaled using predetermined non-integer factors ${ \times1.6, \times2.5, \times3.4, \times4.3 }$. It can be observed that our proposed model effectively captures the patterns of text within the image, and accurately estimates words and numbers with enhanced sharpness and clarity. The enhancement becomes particularly evident in the word 'infinity' in the first row of contexts, 'SUNSHINE' in the fifth row, and the number '0' in the final row, as each of these elements demonstrates considerable improvements.

\subsection{Visualization Local Attention Maps}
\label{subsec::visuallizing_local_attention_maps}

\begin{figure}[t]
    \centering
    \includegraphics[width=1\linewidth]{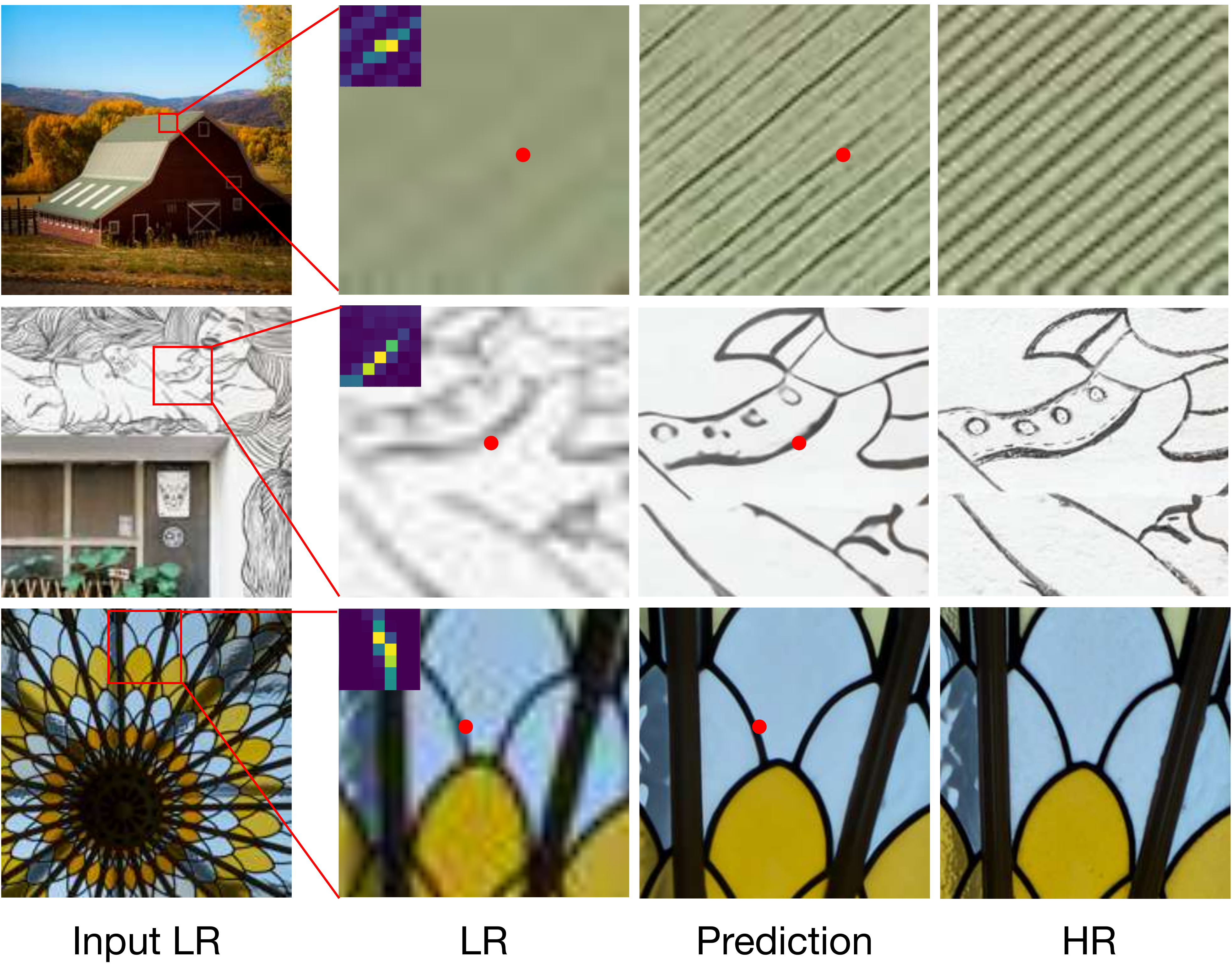}
    \caption{A visualization of local attention maps of the coordinates highlight as \textcolor{red}{red dots}.}
    \label{fig:qualitative_results_attn}
\end{figure}

In Fig.~\ref{fig:qualitative_results_attn}, we providevisualizations of local attention maps. The LR images are generated by applying bicubic downsampling to HR images with scale factors $\{\times 4.5, \times 6, \times 6\}$. Subsequently, the proposed CLIT with RDN \cite{rdn} encoder is utilized to produce HR predictions.  It can be observed that the attention maps closely align with the edges, indicating that the cross-scale local attention block effectively captures similar latent codes within a local area. This substantiates that our proposed design enables the generation of SR images from LR counterparts, ultimately producing clean and sharp edges in the output.

\subsection{Ablation Studies}
\label{subsec::ablation_studies}
In this section, we present a series of ablation analyses to substantiate the design decisions proposed in this paper. All of the ablation experiments are conducted on the DIV2K \cite{div2k} validation set, utilizing the EDSR-baseline \cite{edsr} as the encoder and employing the PSNR metric for evaluation.

\vspace{-10pt}
\paragraph{Validation of the design choices.}
Table~\ref{table:desing_choice} presents a summary of the quantitative contributions associated with each LIT component. A significant improvement is observed when adopting the cross-scale local attention block by comparing LTE with LTE (-a). In addition, LTE (-c) reveals that removing cell decoding leads to more severe performance degradation for in-distribution upsampling scales compared to out-of-distribution counterparts, highlighting the importance of cell decoding for in-distribution upsampling scales. Finally, incorporating the local ensemble technique in LIT (+e) results in a further, albeit modest, enhancement of the overall performance.

\begin{table}[t]
\centering
\footnotesize
\resizebox{1\linewidth}{!}{
\begin{tabular}{l|ccc|ccc}
\multirow{2}{*}{} & \multicolumn{3}{c|}{In-distribution}             & \multicolumn{3}{c}{Out-of-distribution}           \\
                  & $\times 2$     & $\times 3$     & $\times 4$     & $\times 6$     & $\times 8$     & $\times 12$     \\ \hline\hline
LIT               & 34.79          & 31.08          & 29.10          & 26.84          & 25.48          & 23.79           \\ \hline
LIT (-a)          & 34.74          & 31.01          & 29.04          & 26.77          & 25.40          & 23.70           \\
LIT (-c)          & 34.64          & 30.99          & 29.05          & 26.80          & 25.45          & 23.78           \\
LIT (+e)          & \textbf{34.81} & \textbf{31.10} & \textbf{29.12} & \textbf{26.86} & \textbf{25.50} & \textbf{23.81} 
\end{tabular}}
\caption{The average PSNR(dB) of different design choices. -a/c refers to removing cross-scale local attention block, frequency encoding block, and cell size correspondingly. +e refers to adding local ensemble. The \textbf{bolded} indicates the best performance.}
\label{table:desing_choice}
\end{table}
\begin{table}[t]
\centering
\footnotesize
\resizebox{1\linewidth}{!}{
\begin{tabular}{l|ccc|ccc|c}
\multirow{2}{*}{Local gird $G$}         & \multicolumn{3}{c|}{In-distribution}             & \multicolumn{3}{c|}{Out-of-distribution}           &\multirow{2}{*}{Training time}    \\
                                        & $\times 2$     & $\times 3$     & $\times 4$     & $\times 6$     & $\times 8$     & $\times 12$              \\ \hline\hline
$G = 1\times1$                          & 34.74          & 31.01          & 29.04          & 26.77          & 25.40          & 23.70            &22.3   \\
$G = 3\times3$                          & 34.77          & 31.06          & 29.08          & 26.83          & 25.46          & 23.76            &23.5   \\
$G = 5\times5$                          & 34.77          & 31.05          & 29.08          & 26.83          & 25.46          & 23.76            &29.4   \\
$G = 7\times7$                          & 34.79          & 31.08          & 29.10          & 26.84          & 25.48          & \textbf{23.79}   &40.8   \\
$G = 9\times9$                          & 34.81          & 31.09          & 29.11          & 26.85          & \textbf{25.49} & \textbf{23.79}   &59.8   \\
$G = 15\times15$                        & 34.82          & 31.09          & 29.11          & 26.85          & \textbf{25.49} & \textbf{23.79}   &87.0   \\
$G = 25\times25$                        & \textbf{34.83} & \textbf{31.12} & \textbf{29.12} & \textbf{26.86} & 25.48          & 23.78            &211.5
\end{tabular}}
\caption{The average PSNR~(dB) and the training time (hour) of multiple local grid sizes. The \textbf{bolded numbers} correspond to the best performance.}
\label{table:local_grid}
\vspace{-10pt}
\end{table}
\vspace{-10pt}
\paragraph{The effectiveness of the local grid.}
Table \ref{table:local_grid} provides a performance comparison of various local grid sizes employed within the local coordinate sampling scheme, as discussed in Section~\ref{sec::methodology}. The results presented in Table \ref{table:local_grid} reveal that increasing the local grid size contributes to performance enhancements, albeit at the cost of extended training times. As a result, this study adopts a $7\times7$ local grid size to strike an ideal balance between effectiveness and efficiency.

\vspace{-10pt}
\paragraph{Analysis on the training strategy.}
Table~\ref{table:training_strategy} presents a quantitative comparison of the proposed cumulative training strategy with the other training strategies for training an LIT. The baseline strategy trains LIT using upsampling scales sampled from a uniform distribution, $r \sim \mathcal{U}(1, 4)$. Expanding the sampling scale distribution to $r \sim \mathcal{U}(1, 12)$ improves the performance of large-scale upsampling while compromising the performance of small-scale upsampling. To achieve high-quality results across all upsampling scale ranges, the alternative training strategy trains LIT alternatively by switching the sampling scales between $r \sim \mathcal{U}(1, 4)$ and $r \sim \mathcal{U}(4, 12)$. On the other hand, the proposed cumulative training strategy first trains LIT with $r \sim \mathcal{U}(1, 4)$, followed by fine-tuning using the alternative training strategy. This progressive approach improves the performance of LIT by gradually broadening the distribution of the upsampling scales. Comparing the results of the cumulatively trained LIT with the results of EDSR-baseline-CLIT in Table~\ref{table:div2k_results} reveals that the proposed cumulative training strategy works for both LIT and CLIT. Moreover, the results from these tables suggest that the cascaded framework is more effective in addressing arbitrary-scale SR tasks.

To validate the generalizability of the proposed strategy, we apply it on LIIF~\cite{liif} and LTE~\cite{lte} using the same setups, and report the results in Table~\ref{table:cumulative_results}. The results reveal that the cumulative training strategy is able to enhance the performance of LIIF and LTE as well. An interesting fact is that by adopting the proposed cumulative training strategy, the performance of LTE is able to be improved without the need of its original cell clipping operation. The experimental evidence thus justifies the effectiveness and generalizability of our strategy.

\begin{table}[t]
\centering
\footnotesize
\resizebox{1\linewidth}{!}{
\begin{tabular}{l|ccc|ccc}
Training strategy                        & $\times 2$     & $\times 3$     & $\times 4$     & $\times 6$     & $\times 8$     & $\times 12$     \\ 
\hline\hline
Training with $r \sim \mathcal{U}(1, 4)$ & \textbf{34.79} & 31.08          & 29.10          & 26.84          & 25.48          & 23.79           \\
Training with $r \sim \mathcal{U}(1, 12)$& 34.72          & 31.07          & 29.13          & 26.89          & 25.53          & 23.83           \\
Alternative training strategy            & 34.75          & 31.09          & 29.13          & 26.89          & \textbf{25.54} & \textbf{23.84}  \\
Cumulative training strategy             & 34.78          & \textbf{31.11} & \textbf{29.15} & \textbf{26.90} & 25.53          & 23.83
\end{tabular}}
\caption{The average PSNR (dB) of different training strategies. The best performing results are indicated in \textbf{bold}.}
\label{table:training_strategy}
\end{table}

\begin{table}[t]
\centering
\footnotesize
\tabcolsep=1.5pt
\resizebox{1\linewidth}{!}{
\begin{tabular}{l|c|ccccc}
\begin{tabular}[c]{@{}l@{}}Training\\ Strategy\end{tabular}            & Method     & $\times 2$        & $\times 3$        & $\times 4$        & $\times 6$                & $\times 12$      \\ \hline\hline
\multirow{2}{*}{Original}   & EDSR-LIIF~\cite{liif} & 34.67                     & 30.96                     & 29.00                     & 26.75                     & 23.71 \\
                            & EDSR-LTE~\cite{lte}   & 34.72                     & 31.02                     & 29.04                     & 26.81                     & 23.78 \\ \hline
\multirow{2}{*}{Cumulative} & EDSR-LIIF~\cite{liif} & 34.67 (\textbf{+0.00})    & 31.01 (\textbf{+0.05})    & 29.06 (\textbf{+0.06})    & 26.82 (\textbf{+0.07})    & 23.77 (\textbf{+0.06}) \\
                            & EDSR-LTE~\cite{lte}   & 34.72 (\textbf{+0.00})    & 31.04 (\textbf{+0.02})    & 29.09 (\textbf{+0.05})    & 26.86 (\textbf{+0.05})    & 23.81 (\textbf{+0.03})          
\end{tabular}}
\caption{The average PSNR (dB) evaluated on the DIV2K~\cite{div2k} validation set with and without the proposed cumulative training strategy for LIIF~\cite{liif} and LTE~\cite{lte}. The \textbf{bold numbers} correspond to the improvements.}
\label{table:cumulative_results}
\vspace{-10pt}
\end{table}

%%%%%%%%% CONCLUSION
\section{Conclusion}
\label{sec::conclusion}

In this paper, we introduced the attention mechanism and frequency encoding technique to address arbitrary-scale SR tasks. To achieve this objective, we proposed the LIT framework, which consists of a cross-scale local attention block and a local frequency encoding block. The former is designed to find the local latent embedding to reconstruct the corresponding RGB value, while the latter focuses on encoding coordinates with the frequency information derived from the feature embeddings. To enhance the capability of capturing fine details, we proposed CLIT and the corresponding cumulative training strategy that trains the model with progressively increasing upsampling scales. Based on the experimental results, both quantitative and qualitative assessments showcase the superior performance of LIT and CLIT in comparison to existing methods. Furthermore, our comprehensive analyses validated the effectiveness of the training strategies and the components employed by them.

%%%%%%%%% Acknowledgement
\section*{Acknowledgements}

The authors gratefully acknowledge the support from the National Science and Technology Council (NSTC) in Taiwan under grant numbers MOST 111-2223-E-007-004-MY3 and MOST 111-2628-E-007-010, as well as the financial support from MediaTek Inc., Taiwan.

%------------------------------------------------------------------------

%%%%%%%%% REFERENCES
{\small
\bibliographystyle{mycvpretal}
\bibliography{egbib}

\newcommand{\IJCV}{Int. J. Computer Vision (IJCV)}\newcommand{\CVPR}{Proc. IEEE
  Conf. on Computer Vision and Pattern Recognition
  (CVPR)}\newcommand{\CVPRW}{Proc. IEEE Conf. on Computer Vision and Pattern
  Recognition Workshop (CVPRW)}\newcommand{\ICCV}{Proc. IEEE Int. Conf. on
  Computer Vision (ICCV)}\newcommand{\ICCVW}{Proc. IEEE Int. Conf. on Computer
  Vision Workshop (ICCVW)}\newcommand{\ECCV}{Proc. European Conf. on Computer
  Vision (ECCV)}\newcommand{\ECCVW}{Proc. European Conf. on Computer Vision
  Workshop (ECCVW)}\newcommand{\IROS}{Proc. IEEE Int. Conf. on Intelligent
  Robots and Systems (IROS)}\newcommand{\CoRL}{Proc. Conf. on Robot Learning
  (CoRL)}\newcommand{\ICRA}{Proc. IEEE Int. Conf. on Robotics and Automation
  (ICRA)}\newcommand{\AAAI}{Proc. AAAI Conf. on Artificial Intelligence
  (AAAI)}\newcommand{\IJCAI}{Proc. Int. Joint Conf. on Artificial Intelligence
  (IJCAI)}\newcommand{\PAMI}{IEEE Trans. Pattern Analysis and Machine
  Intelligence (TPAMI)}\newcommand{\NIPS}{Proc. Conf. on Neural Information
  Processing Systems (NeurIPS)}\newcommand{\ICML}{Proc. Int. Conf. on Machine
  Learning (ICML)}\newcommand{\ICLR}{Proc. Int. Conf. on Learning
  Representations (ICLR)}\newcommand{\ICLRW}{Proc. Int. Conf. on Learning
  Representations Workshop (ICLRW)}\newcommand{\ICASSP}{Proc. IEEE Int. Conf.
  on Acoustics, Speech, & Signal Processing (ICASSP)}\newcommand{\BMVC}{Proc.
  British Machine Vision Conf. (BMVC)}\newcommand{\ACCV}{Proc. Asian Conf. on
  Computer Vision (ACCV)}\newcommand{\WACV}{Proc. IEEE Winter Conf. on
  Applications of Computer Vision (WACV)}\newcommand{\NAACL}{Proc. Conf. of
  North American Chapter of the Association for Computational Linguistics:
  Human Language Technologies (NAACL)}
\begin{thebibliography}{51}
\providecommand{\natexlab}[1]{#1}
\providecommand{\url}[1]{\texttt{#1}}
\expandafter\ifx\csname urlstyle\endcsname\relax
  \providecommand{\doi}[1]{doi: #1}\else
  \providecommand{\doi}{doi: \begingroup \urlstyle{rm}\Url}\fi

\bibitem[Dong et~al.(2016)Dong, Loy, He, and Tang]{srcnn}
C.~Dong, C.~C. Loy, K.~He, and X.~Tang.
\newblock Image super-resolution using deep convolutional networks.
\newblock \emph{\PAMI}, 38\penalty0 (2):\penalty0 295--307, 2016.

\bibitem[Kim et~al.(2016)Kim, Lee, and Lee]{vdsr}
J.~Kim, J.~K. Lee, and K.~M. Lee.
\newblock Accurate image super-resolution using very deep convolutional
  networks.
\newblock In \emph{\CVPR}, pages 1646--1654, 2016.

\bibitem[Dong et~al.(2016)Dong, Loy, and Tang]{fsrcnn}
C.~Dong, C.~C. Loy, and X.~Tang.
\newblock Accelerating the super-resolution convolutional neural network.
\newblock In \emph{\ECCV}, pages 391--407, 2016.

\bibitem[Shi et~al.(2016)Shi, Caballero, Huszar, et~al.]{espcn}
W.~Shi, J.~Caballero, F.~Huszar, J.~Totz, A.~P. Aitken, et~al.
\newblock Real-time single image and video super-resolution using an efficient
  sub-pixel convolutional neural network.
\newblock In \emph{\CVPR}, pages 1874--1883, 2016.

\bibitem[Lai et~al.(2019)Lai, Huang, Ahuja, and Yang]{lapsrn}
W.~Lai, J.~Huang, N.~Ahuja, and M.~Yang.
\newblock Fast and accurate image super-resolution with deep laplacian pyramid
  networks.
\newblock \emph{\PAMI}, 41\penalty0 (11):\penalty0 2599--2613, 2019.

\bibitem[Ledig et~al.(2017)Ledig, Theis, Huszar, et~al.]{srresnet}
C.~Ledig, L.~Theis, F.~Huszar, J.~Caballero, A.~Cunningham, et~al.
\newblock Photo-realistic single image super-resolution using a generative
  adversarial network.
\newblock In \emph{\CVPR}, pages 105--114, 2017.

\bibitem[Lim et~al.(2017)Lim, Son, Kim, et~al.]{edsr}
B.~Lim, S.~Son, H.~Kim, S.~Nah, and K.~M. Lee.
\newblock Enhanced deep residual networks for single image super-resolution.
\newblock In \emph{\CVPRW}, pages 1132--1140, 2017.

\bibitem[Wang et~al.(2018)Wang, Yu, Wu, et~al.]{esrgan}
X.~Wang, K.~Yu, S.~Wu, J.~Gu, Y.~Liu, et~al.
\newblock Esrgan: Enhanced super-resolution generative adversarial networks.
\newblock In \emph{\ECCVW}, pages 63--79, 2018.

\bibitem[Zhang et~al.(2018{\natexlab{a}})Zhang, Tian, Kong, et~al.]{rdn}
Y.~Zhang, Y.~Tian, Y.~Kong, B.~Zhong, and Y.~Fu.
\newblock Residual dense network for image super-resolution.
\newblock In \emph{\CVPR}, pages 2472--2481, 2018{\natexlab{a}}.

\bibitem[Zhang et~al.(2018{\natexlab{b}})Zhang, Li, Li, et~al.]{rcan}
Y.~Zhang, K.~Li, K.~Li, L.~Wang, B.~Zhong, et~al.
\newblock Image super-resolution using very deep residual channel attention
  networks.
\newblock In \emph{\ECCV}, pages 294--310, 2018{\natexlab{b}}.

\bibitem[Dai et~al.(2019)Dai, Cai, Zhang, et~al.]{san}
T.~Dai, J.~Cai, Y.~Zhang, S.~Xia, and L.~Zhang.
\newblock Second-order attention network for single image super-resolution.
\newblock In \emph{\CVPR}, pages 11065--11074, 2019.

\bibitem[Mei et~al.(2020)Mei, Fan, Zhou, et~al.]{csnl}
Y.~Mei, Y.~Fan, Y.~Zhou, L.~Huang, T.~S. Huang, et~al.
\newblock Image super-resolution with cross-scale non-local attention and
  exhaustive self-exemplars mining.
\newblock In \emph{\CVPR}, pages 5689--5698, 2020.

\bibitem[Chen et~al.(2021)Chen, Wang, Guo, et~al.]{ipt}
H.~Chen, Y.~Wang, T.~Guo, C.~Xu, Y.~Deng, et~al.
\newblock Pre-trained image processing transformer.
\newblock In \emph{\CVPR}, pages 12294--12305, 2021.

\bibitem[Liang et~al.(2021)Liang, Cao, Sun, et~al.]{swinir}
J.~Liang, J.~Cao, G.~Sun, K.~Zhang, L.~Van~Gool, et~al.
\newblock Swinir: Image restoration using swin transformer.
\newblock In \emph{\ICCVW}, pages 1833--1844, 2021.

\bibitem[Chen et~al.(2022)Chen, Wang, Zhou, and Dong]{hat}
X.~Chen, X.~Wang, J.~Zhou, and C.~Dong.
\newblock Activating more pixels in image super-resolution transformer.
\newblock \emph{CoRR}, abs/2205.04437, 2022.

\bibitem[Hu et~al.(2019)Hu, Mu, Zhang, et~al.]{metasr}
X.~Hu, H.~Mu, X.~Zhang, Z.~Wang, T.~Tan, et~al.
\newblock Meta-sr: {A} magnification-arbitrary network for super-resolution.
\newblock In \emph{\CVPR}, pages 1575--1584, 2019.

\bibitem[Chen et~al.(2021)Chen, Liu, and Wang]{liif}
Y.~Chen, S.~Liu, and X.~Wang.
\newblock Learning continuous image representation with local implicit image
  function.
\newblock In \emph{\CVPR}, pages 8628--8638, 2021.

\bibitem[Xu et~al.(2021)Xu, Wang, and Shi]{ultrasr}
X.~Xu, Z.~Wang, and H.~Shi.
\newblock Ultrasr: Spatial encoding is a missing key for implicit image
  function-based arbitrary-scale super-resolution.
\newblock \emph{CoRR}, abs/2103.12716, 2021.

\bibitem[Liu et~al.(2021)Liu, Guo, and Zhang]{ipe}
Y.~Liu, Y.~Guo, and S.~Zhang.
\newblock Enhancing multi-scale implicit learning in image super-resolution
  with integrated positional encoding.
\newblock \emph{CoRR}, abs/2112.05756, 2021.

\bibitem[Yang et~al.(2021)Yang, Shen, Yue, and Li]{itsrn}
J.~Yang, S.~Shen, H.~Yue, and K.~Li.
\newblock Implicit transformer network for screen content image continuous
  super-resolution.
\newblock In \emph{\NIPS}, pages 13304--13315, 2021.

\bibitem[Lee and Jin(2022)]{lte}
J.~Lee and K.~H. Jin.
\newblock Local texture estimator for implicit representation function.
\newblock In \emph{\CVPR}, pages 1929--1938, 2022.

\bibitem[Vaswani et~al.(2017)Vaswani, Shazeer, Parmar, et~al.]{transformer}
A.~Vaswani, N.~Shazeer, N.~Parmar, J.~Uszkoreit, L.~Jones, et~al.
\newblock Attention is all you need.
\newblock In \emph{\NIPS}, pages 5998--6008, 2017.

\bibitem[Liu et~al.(2021)Liu, Lin, Cao, et~al.]{swin}
Z.~Liu, Y.~Lin, Y.~Cao, H.~Hu, Y.~Wei, et~al.
\newblock Swin transformer: Hierarchical vision transformer using shifted
  windows.
\newblock In \emph{\ICCV}, 2021.

\bibitem[Wang et~al.(2022)Wang, Yao, Chen, et~al.]{crossformer}
W.~Wang, L.~Yao, L.~Chen, B.~Lin, D.~Cai, et~al.
\newblock Crossformer: {A} versatile vision transformer hinging on cross-scale
  attention.
\newblock In \emph{\ICLR}, 2022.

\bibitem[Chen and Zhang(2019)]{chen2019}
Z.~Chen and H.~Zhang.
\newblock Learning implicit fields for generative shape modeling.
\newblock In \emph{\CVPR}, pages 5939--5948, 2019.

\bibitem[Park et~al.(2019)Park, Florence, Straub, et~al.]{deepsdf}
J.~J. Park, P.~Florence, J.~Straub, R.~A. Newcombe, and S.~Lovegrove.
\newblock Deepsdf: Learning continuous signed distance functions for shape
  representation.
\newblock In \emph{\CVPR}, pages 165--174, 2019.

\bibitem[Mescheder et~al.(2019)Mescheder, Oechsle, Niemeyer,
  et~al.]{mescheder2019}
L.~M. Mescheder, M.~Oechsle, M.~Niemeyer, S.~Nowozin, and A.~Geiger.
\newblock Occupancy networks: Learning 3d reconstruction in function space.
\newblock In \emph{\CVPR}, pages 4460--4470, 2019.

\bibitem[Genova et~al.(2020)Genova, Cole, Sud, et~al.]{genova2020}
K.~Genova, F.~Cole, A.~Sud, A.~Sarna, and T.~A. Funkhouser.
\newblock Local deep implicit functions for 3d shape.
\newblock In \emph{\CVPR}, pages 4856--4865, 2020.

\bibitem[Atzmon and Lipman(2020)]{sal}
M.~Atzmon and Y.~Lipman.
\newblock {SAL:} sign agnostic learning of shapes from raw data.
\newblock In \emph{\CVPR}, pages 2562--2571, 2020.

\bibitem[Sitzmann et~al.(2019)Sitzmann, Zollh{\"{o}}fer, and Wetzstein]{srn}
V.~Sitzmann, M.~Zollh{\"{o}}fer, and G.~Wetzstein.
\newblock Scene representation networks: Continuous 3d-structure-aware neural
  scene representations.
\newblock In \emph{\NIPS}, pages 1119--1130, 2019.

\bibitem[Jiang et~al.(2020)Jiang, Sud, Makadia, et~al.]{jiang2020}
C.~M. Jiang, A.~Sud, A.~Makadia, J.~Huang, M.~Nie{\ss}ner, et~al.
\newblock Local implicit grid representations for 3d scenes.
\newblock In \emph{\CVPR}, pages 6000--6009, 2020.

\bibitem[Chabra et~al.(2020)Chabra, Lenssen, Ilg, et~al.]{chabra2020}
R.~Chabra, J.~E. Lenssen, E.~Ilg, T.~Schmidt, J.~Straub, et~al.
\newblock Deep local shapes: Learning local {SDF} priors for detailed 3d
  reconstruction.
\newblock In \emph{\ECCV}, pages 608--625, 2020.

\bibitem[Peng et~al.(2020)Peng, Niemeyer, Mescheder, et~al.]{peng2020}
S.~Peng, M.~Niemeyer, L.~M. Mescheder, M.~Pollefeys, and A.~Geiger.
\newblock Convolutional occupancy networks.
\newblock In \emph{\ECCV}, pages 523--540, 2020.

\bibitem[Mildenhall et~al.(2020)Mildenhall, Srinivasan, Tancik, et~al.]{nerf}
B.~Mildenhall, P.~P. Srinivasan, M.~Tancik, J.~T. Barron, R.~Ramamoorthi,
  et~al.
\newblock Nerf: Representing scenes as neural radiance fields for view
  synthesis.
\newblock In \emph{\ECCV}, pages 405--421, 2020.

\bibitem[Niemeyer et~al.(2020)Niemeyer, Mescheder, Oechsle, and
  Geiger]{niemeyer2020}
M.~Niemeyer, L.~Mescheder, M.~Oechsle, and A.~Geiger.
\newblock Differentiable volumetric rendering: Learning implicit 3d
  representations without 3d supervision.
\newblock In \emph{\CVPR}, pages 3501--3512, 2020.

\bibitem[Liu et~al.(2020)Liu, Gu, Lin, et~al.]{nsvf}
L.~Liu, J.~Gu, K.~Z. Lin, T.~Chua, and C.~Theobalt.
\newblock Neural sparse voxel fields.
\newblock In \emph{\NIPS}, 2020.

\bibitem[Barron et~al.(2021)Barron, Mildenhall, Tancik, et~al.]{mipnerf}
J.~T. Barron, B.~Mildenhall, M.~Tancik, P.~Hedman, R.~Martin{-}Brualla, et~al.
\newblock Mip-nerf: {A} multiscale representation for anti-aliasing neural
  radiance fields.
\newblock In \emph{\ICCV}, pages 5835--5844, 2021.

\bibitem[Klocek et~al.(2019)Klocek, Maziarka, Wo{\l}czyk, et~al.]{klocek2019}
S.~Klocek, {\L}.~Maziarka, M.~Wo{\l}czyk, J.~Tabor, J.~Nowak, et~al.
\newblock Hypernetwork functional image representation.
\newblock In \emph{Artificial Neural Networks and Machine Learning}, pages
  496--510, 2019.

\bibitem[Sitzmann et~al.(2020)Sitzmann, Martel, Bergman, et~al.]{siren}
V.~Sitzmann, J.~N.~P. Martel, A.~W. Bergman, D.~B. Lindell, and G.~Wetzstein.
\newblock Implicit neural representations with periodic activation functions.
\newblock In \emph{\NIPS}, pages 7462--7473, 2020.

\bibitem[Grm et~al.(2020)Grm, Scheirer, and Struc]{csrip}
K.~Grm, W.~J. Scheirer, and V.~Struc.
\newblock Face hallucination using cascaded super-resolution and identity
  priors.
\newblock \emph{{IEEE} Trans. Image Processing}, pages 2150--2165, 2020.

\bibitem[Fu et~al.(2019)Fu, Kong, Zheng, et~al.]{cdpn}
Z.~Fu, Y.~Kong, Y.~Zheng, H.~Ye, W.~Hu, et~al.
\newblock Cascaded detail-preserving networks for super-resolution of document
  images.
\newblock In \emph{Int. Conf. on Doc. Analysis Recognition, {ICDAR}}, pages
  240--245, 2019.

\bibitem[Rahaman et~al.(2019)Rahaman, Baratin, Arpit, et~al.]{rahaman2019}
N.~Rahaman, A.~Baratin, D.~Arpit, F.~Draxler, M.~Lin, et~al.
\newblock On the spectral bias of neural networks.
\newblock In \emph{\ICML}, pages 5301--5310, 2019.

\bibitem[Basri et~al.(2020)Basri, Galun, Geifman, et~al.]{basri2020}
R.~Basri, M.~Galun, A.~Geifman, D.~W. Jacobs, Y.~Kasten, et~al.
\newblock Frequency bias in neural networks for input of non-uniform density.
\newblock In \emph{\ICML}, pages 685--694, 2020.

\bibitem[Tancik et~al.(2021)Tancik, Mildenhall, Wang, et~al.]{tancik2021}
M.~Tancik, B.~Mildenhall, T.~Wang, D.~Schmidt, P.~P. Srinivasan, et~al.
\newblock Learned initializations for optimizing coordinate-based neural
  representations.
\newblock In \emph{\CVPR}, pages 2846--2855, 2021.

\bibitem[Hendrycks and Gimpel(2016)]{gelu}
D.~Hendrycks and K.~Gimpel.
\newblock Bridging nonlinearities and stochastic regularizers with gaussian
  error linear units.
\newblock \emph{CoRR}, abs/1606.08415, 2016.

\bibitem[Agustsson and Timofte(2017)]{div2k}
E.~Agustsson and R.~Timofte.
\newblock {NTIRE} 2017 challenge on single image super-resolution: Dataset and
  study.
\newblock In \emph{\CVPRW}, pages 1122--1131, 2017.

\bibitem[Bevilacqua et~al.(2012)Bevilacqua, Roumy, Guillemot, and
  Alberi{-}Morel]{set5}
M.~Bevilacqua, A.~Roumy, C.~Guillemot, and M.~Alberi{-}Morel.
\newblock Low-complexity single-image super-resolution based on nonnegative
  neighbor embedding.
\newblock In \emph{\BMVC}, pages 1--10, 2012.

\bibitem[Zeyde et~al.(2010)Zeyde, Elad, and Protter]{set14}
R.~Zeyde, M.~Elad, and M.~Protter.
\newblock On single image scale-up using sparse-representations.
\newblock In \emph{Curves and Surfaces}, volume 6920 of \emph{Lecture Notes in
  Computer Science}, pages 711--730, 2010.

\bibitem[Martin et~al.(2001)Martin, Fowlkes, Tal, and Malik]{b100}
D.~R. Martin, C.~C. Fowlkes, D.~Tal, and J.~Malik.
\newblock A database of human segmented natural images and its application to
  evaluating segmentation algorithms and measuring ecological statistics.
\newblock In \emph{\ICCV}, pages 416--425, 2001.

\bibitem[Huang et~al.(2015)Huang, Singh, and Ahuja]{urban100}
J.~Huang, A.~Singh, and N.~Ahuja.
\newblock Single image super-resolution from transformed self-exemplars.
\newblock In \emph{\CVPR}, pages 5197--5206, 2015.

\bibitem[Kingma and Ba(2015)]{adam}
D.~P. Kingma and J.~Ba.
\newblock Adam: A method for stochastic optimization.
\newblock In \emph{\ICLR}, 2015.

\end{thebibliography}
}

\end{document}